\documentclass[11pt]{article}

\usepackage[preprint]{acl}

\usepackage{times}
\usepackage{latexsym}

\usepackage[T1]{fontenc}

\usepackage[utf8]{inputenc}

\usepackage{microtype}

\usepackage{inconsolata}

\usepackage{graphicx}
\usepackage{subcaption}
\usepackage{amsmath}
\usepackage{amssymb}
\usepackage{booktabs}

\title{Boosting LLM Exploration via Weak-Model Guidance in RLVR}

\author{
 \textbf{Xingyu Shen\textsuperscript{1}},
 \textbf{Huishuai Zhang\textsuperscript{1}},
 \textbf{Peng Li},
 \textbf{Yinchun Wang},
 \textbf{Dongyan Zhao\textsuperscript{1,2}}\thanks{Corresponding author.},
\\
\\
 \textsuperscript{1}Wangxuan Institute of Computer Technology, Peking University,\\
 \textsuperscript{2}National Engineering Research Center of New Electronic Publishing Technologies
\\
\texttt{\{shenxy,zhanghuishuai,zhaody\}@pku.edu.cn}
\\
}

\usepackage[most]{tcolorbox}
\usepackage{xcolor}
\usepackage{times} 
\definecolor{border}{HTML}{4A90D2}  
\definecolor{bg}{HTML}{EBF3FB}      
\definecolor{q}{HTML}{4A90D2}       

\newtcolorbox{promptbox}[1]{
    colback=bg,
    colframe=border,
    title=#1,
    fonttitle=\large\fontfamily{ptm}\selectfont\textbf,
    boxrule=0.8pt,
    arc=2mm,
    left=2mm,
    right=2mm,
    top=1mm,
    bottom=1mm,
    fontupper=\fontfamily{ptm}\selectfont
}

\begin{document}
\maketitle

\begin{abstract}

Reinforcement Learning with Verifiable Rewards (RLVR) significantly improves LLM reasoning but often causes a drop in policy entropy, leading to narrowed reasoning coverage and degraded pass@$k$ for large $k$. While existing methods mitigate this entropy collapse through algorithmic regularizations, cross-model non-parametric perturbation is also neglected. In this work, we propose a simple yet effective approach to preserve the generative diversity of LLMs during RLVR. Instead of relying solely on internal exploration, we force the target model to generate answers based on partial reasoning trajectories generated by a smaller, weaker language models. These unfamiliar prefixes effectively disrupt over-confidence and encourage the exploration of distinct reasoning paths. We empirically study the potential of outer prefixes, revealing the mechanism of the impact of distributional discrepancy to the exploration dynamics in RLVR training. Experiments across multiple mathematical benchmarks show that our method consistently outperforms vanilla RLVR. Notably, the performance gain becomes increasingly pronounced as $k$ scales up, demonstrating a substantial expansion of reasoning coverage. Furthermore, our approach efficiently mitigates entropy collapse without requiring additional SFT, intricate reward designs, or complex prompting.

\end{abstract}

\section{Introduction}

Reinforcement Learning with Verifiable Rewards (RLVR) has been widely adopted in the post-training stage for eliciting strong reasoning from Large Language Models (LLMs). Given an input prompt, the model generates a response and receives feedback from an external verifier. The training objective increases the probability of generating high-reward responses while penalizing low-reward ones, thereby enabling sequence-level optimization without requiring additional human-curated annotations \citep{schulman2017proximal, shao2024deepseekmath}. RLVR has been extensively used in both open-source and closed-source LLMs to improve their ability to solve challenging reasoning tasks, as demonstrated by models such as DeepSeek-R1 \citep{guo2025deepseek}, GPT-5 \citep{singh2025openai}, and Gemini \citep{comanici2025gemini}. These models have achieved high accuracy on various mathematical reasoning benchmarks. However, most existing evaluations focus primarily on pass@1 performance, namely the accuracy achieved with a single trial, while largely overlooking reasoning coverage.

During RLVR training, the model is optimized to maximize the probability of generating reasoning trajectories that lead to correct answers. \citet{cui2025entropy} point out that LLMs tend to trade entropy for performance during reinforcement learning. Specifically, they observe that policy entropy decreases sharply in the early stage of training, which limits the model's ability to conduct broad exploration. As a result, once the model becomes over-confident after a few training steps, it may have fewer opportunities to discover alternative reasoning paths that could also lead to correct answers and subsequently increase their probabilities. Along similar lines, \citet{yue2025does} find that RLVR primarily increases the model's confidence in reasoning paths that have already been established, while narrowing the overall reasoning coverage. They support this conclusion using pass@$k$ metrics \citep{brown2024large}, where a problem is considered solved if the model produces a correct answer within $k$ sampled responses. When $k$ is sufficiently large, the base model can outperform the RLVR-trained model, suggesting that RLVR may improve single-sample accuracy at the cost of reduced solution diversity.

Many approaches have been proposed to alleviate the degradation of pass@$k$ performance of RLVR. \citet{cui2025entropy} attempt to broaden exploration by incorporating entropy regularization into the training process, thereby mitigating over-confidence in a limited set of reasoning paths. \citet{peng2025simko} seek to control training dynamics by calibrating the policy gradients of selected tokens. Other works focus on reward and advantage-function design to encourage exploration during training \citep{dong2025rl, chen2025pass}. While these methods modify internal training objectives or redistribute gradients, they still operate within the target model's inherent search space. In contrast, our work introduces an orthogonal perspective by leveraging cross-model generational diversity.

In this work, we first investigate the differences among generations produced by different language models, revealing distributional gaps between the outputs of larger and smaller models. Due to differences in pretraining data, model architecture, and optimization dynamics, different models may produce substantially distinct reasoning trajectories for the same problem. This observation motivates a simple hypothesis: if the target model is guided to continue from unfamiliar partial reasoning trajectories, it may be encouraged to explore alternative solution paths that would otherwise be assigned low probability. Notably, our approach differs from knowledge distillation, where stronger models are typically used to generate high-quality responses for training weaker models \citep{park2019relational, taori2023stanford}. Instead, we use smaller and weaker base models to generate partial reasoning trajectories. These trajectories may not directly lead to optimal solutions, but we hypothesize that they can serve as non-parametric steering signals, providing the target model with diverse initial reasoning states and forcing the policy to explore paths that may be ignored under standard RLVR due to over-confidence.

We evaluate our method against a standard RLVR baseline in which the model generates responses solely from the original question, without any reasoning-path prefix. Experiments on multiple models and mathematical reasoning benchmarks show that our method outperforms the vanilla RLVR-trained model in most settings. Moreover, the improvement in pass@$k$ becomes more pronounced as $k$ increases, suggesting that our approach can effectively mitigate GRPO-induced diversity collapse and improve reasoning coverage.

    
    

Our main contributions are as follows:
\begin{itemize}
    \item We provide a systematic study of behavioral variation across language models and show that cross-model generational diversity has a significant impact on the exploration dynamics of RLVR.

    \item We introduce a prefix-guided RLVR framework that uses partial reasoning trajectories from auxiliary models to steer the target model toward under-explored reasoning regions. Across multiple reasoning benchmarks, our method consistently improves over vanilla GRPO. The gains become more pronounced for larger pass@$k$, demonstrating that our approach effectively improves reasoning coverage and alleviates diversity collapse. 

    \item We perform extensive analyses to understand why small auxiliary models can enhance the RLVR training of larger models. Our results suggest that cross-model prefix guidance acts as a simple yet effective mechanism for mitigating entropy collapse, even when the auxiliary prefixes provide little or occasionally misleading guidance.
\end{itemize}

\section{Preliminaries on GRPO and Entropy}

\paragraph{Group Relative Policy Optimization (GRPO).}
GRPO \citep{shao2024deepseekmath} is a reinforcement learning algorithm adapted from Proximal Policy Optimization (PPO) \citep{schulman2017proximal}. It is widely used in RLVR frameworks because it avoids training a separate value model and instead estimates advantages by comparing multiple sampled responses to the same prompt.

Let \(\pi_{\theta}\) denote the policy model parameterized by \(\theta\), and let \(\pi_{\theta_{\mathrm{old}}}\) denote the policy used to generate rollouts before the current update. Given a prompt \(q \), GRPO samples a group of \(G\) responses
\[
    \{r_i\}_{i=1}^{G} \sim \pi_{\theta_{\mathrm{old}}}(\cdot \mid q).
\]
Each response \(r_i\) is evaluated by a verifier or reward model, producing a scalar reward \(R_i\). GRPO then computes a group-normalized advantage for each sampled output:
\begin{equation}
    \hat{A}_{i}
    =
    \frac{
        R_i - \operatorname{mean}(\{R_j\}_{j=1}^{G})
    }{
        \operatorname{std}(\{R_j\}_{j=1}^{G})
    }.
\end{equation}
The same sequence-level advantage \(\hat{A}_i\) is assigned to all tokens in the corresponding response \(r_i\).

The GRPO objective adopts a PPO-style clipped surrogate loss with a KL penalty against a frozen reference model \(\pi_{\mathrm{ref}}\). Specifically, the objective is defined as
\begin{equation*}
\begin{split}
&\mathcal{J}_{\mathrm{GRPO}}(\theta)
=
\mathbb{E}_{q, \{r_i\}_{i=1}^{G} \sim \pi_{\theta_{\mathrm{old}}}}
\Bigg[
\frac{1}{G}
\sum_{i=1}^{G}
\frac{1}{|r_i|}
\sum_{k=1}^{|r_i|}
\Bigg(\\
&\quad\min \Big(
\rho_{i,k}(\theta)\hat{A}_{i},
\operatorname{clip}\big(\rho_{i,k}(\theta), 1-\epsilon, 1+\epsilon\big)\hat{A}_{i}
\Big)
\\
&\quad 
- \beta \,\mathbb{D}_{\mathrm{KL}}\big(\pi_{\theta}\parallel \pi_{\mathrm{ref}}\big)
\Bigg)
\Bigg],
\end{split}
\end{equation*}
where \(\rho_{i,k}(\theta)=
    \frac{
        \pi_{\theta}(r_{i,k} \mid q, r_{i,<k})
    }{
        \pi_{\theta_{\mathrm{old}}}(r_{i,k} \mid q, r_{i,<k})
    }\)
is the token-level importance sampling ratio, \(\epsilon\) is the clipping coefficient, and \(\beta\) controls the strength of the KL regularization.





\paragraph{Entropy Collapse in RLVR.}
The training dynamics of GRPO reveal an inherent trade-off between exploitation and exploration. As the model increases the probability of generating high-reward responses, it also assigns greater confidence to its own preferred reasoning trajectories, often at the cost of generation diversity. Consequently, under repeated sampling, the model may solve certain queries more frequently, while for other queries, its probability mass may collapse onto erroneous reasoning paths. This phenomenon is closely related to entropy collapse, where the entropy of the policy decreases rapidly as training progresses \citep{cui2025entropy, yu2026dapo}.

Formally, we define the token-level policy entropy at position \(k\) as the uncertainty of the policy distribution after observing the preceding tokens:

{\small
\begin{equation}
    H_{\theta}(t_k \mid t_{<k})
    =
    - \sum_{t_w \in \mathcal{V}}
    \pi_{\theta}(t_w \mid t_{<k})
    \log \pi_{\theta}(t_w \mid t_{<k}),\nonumber
\end{equation}}
where \(\mathcal{V}\) denotes the vocabulary of the policy model. When the policy becomes highly confident in a particular token \(t_w\), i.e., when \(\pi_{\theta}(t_w \mid t_{<k})\) approaches \(1\), the entropy approaches \(0\), indicating reduced uncertainty and weaker exploration.

To enable a more fine-grained analysis of entropy dynamics, we further segment each reasoning trajectory into a sequence of reasoning steps according to newline characters and periods. Specifically, we denote a reasoning trajectory as \(r = \{s_1, \cdots, s_m\},\) where each step \(s_j = \{t_1, \cdots, t_l\}\) consists of a sequence of tokens. We then define the step-level entropy as the average token-level entropy within that step:
{ 
\begin{equation}
    \bar{H}_{\theta}(s_j)
    =
    \frac{1}{l}
    \sum_{k=1}^{l}
    H_{\theta}(t_k \mid s_{<j}, t_{<k}),
\end{equation}}
where \(s_{<j}\) denotes the preceding reasoning steps and \(t_{<k}\) denotes the preceding tokens within the current step. Here, \(\theta\) refers to the parameters of the target policy model being trained, rather than those of the auxiliary small model used to generate prefix trajectories.

\section{Motivation: Cross-Model Prefixes Increase Policy Uncertainty}
\label{sec:cross_model_entropy}

To understand how reasoning trajectories generated by weaker auxiliary models can influence the target model, we first empirically investigate the entropy discrepancy induced by cross-model prefixes.

We observe that when the target model is conditioned on token or step sequences generated by another model, it exhibits weaker adaptability to autonomous continuation than when it is conditioned on its own generated content. In particular, the target model becomes more uncertain when selecting subsequent tokens, which is reflected by higher policy entropy. This suggests that cross-model prefixes may shift the target model away from its familiar generation distribution and expose it to under-explored regions of the reasoning space.

\begin{figure}[!htbp]
    \centering
    \includegraphics[width=\columnwidth]{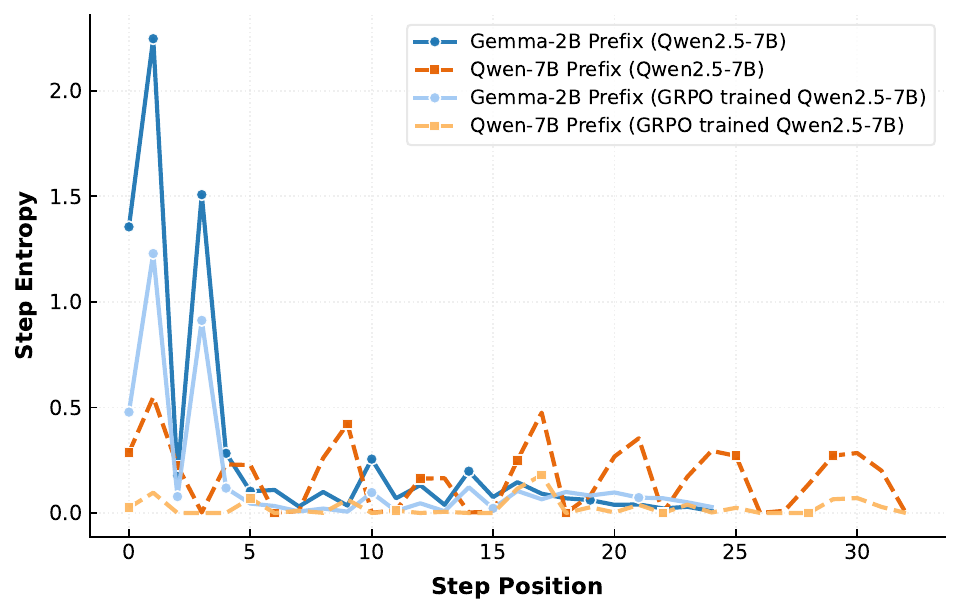}
    \vspace{-0.5em} 
    \caption{Step-level entropy dynamics of Qwen2.5-7B conditioned on prefixes generated by Gemma-2-2B and Qwen2.5-7B itself, before and after GRPO training. Cross-model prefixes induce higher entropy in the early reasoning steps, suggesting broader exploration.}
    \label{fig:step_entropy_dynamics}
\end{figure}

To illustrate this phenomenon, we present a representative example in which we compute the step-level entropy of Qwen2.5-7B before and after GRPO training. We compare two types of prefixes: prefixes generated by Gemma-2-2B and prefixes generated by Qwen2.5-7B itself. The results are shown in Figure~\ref{fig:step_entropy_dynamics}. In the initial reasoning steps, Qwen2.5-7B exhibits substantially higher entropy when conditioned on the Gemma-2-2B-generated prefix than when conditioned on its own generated prefix. This indicates that the target model is less certain about how to continue from reasoning states produced by the smaller model.

As generation proceeds, the step-level entropy gradually decreases and eventually approaches a similar level across the two types of prefixes. This suggests that the target model can adapt to the unfamiliar prefix after several reasoning steps, but the early-stage uncertainty provides an opportunity to broaden the exploration space. When the same analysis is conducted using the GRPO-trained model, the entropy values decrease overall, consistent with the entropy-collapse behavior of RLVR. Nevertheless, the relative trend remains similar: cross-model prefixes still induce higher uncertainty in the early reasoning steps.

This observation motivates our framework. By injecting partial reasoning trajectories generated by auxiliary models into the RLVR process, we can perturb the target model away from its over-confident reasoning paths and encourage exploration of alternative trajectories. In this way, cross-model prefixes serve as non-parametric steering signals that help mitigate diversity collapse during RLVR training.

\section{Methodology}

\begin{figure*}[t]
    \centering
    \includegraphics[width=0.9\textwidth, trim=1cm 1.5cm 1cm 0cm, clip]{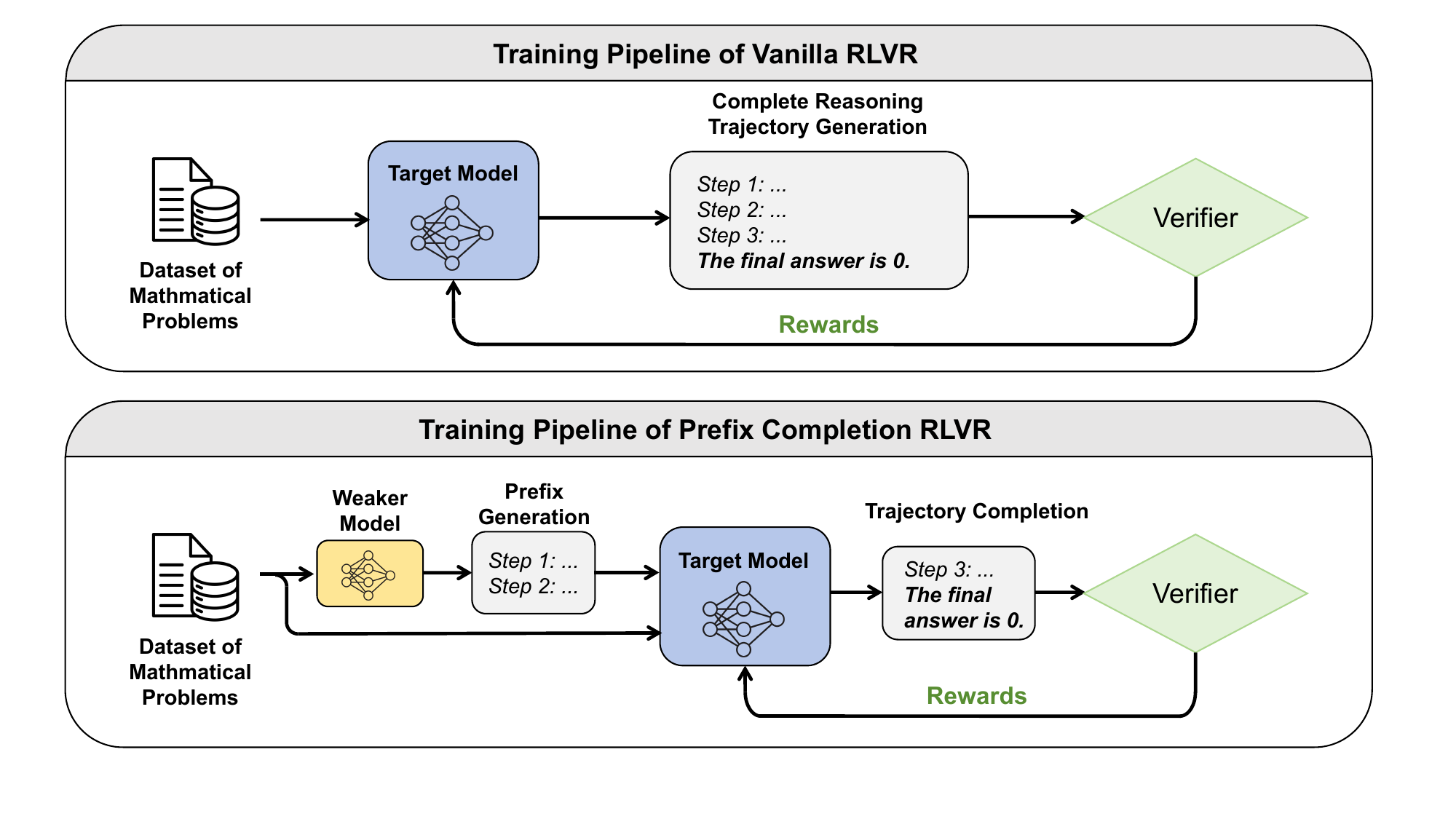}
    \caption{Overview of the proposed prefix-completion RLVR framework. Auxiliary small models first generate candidate reasoning trajectories, which are then truncated into partial prefixes. During RLVR training, the target model is trained to complete the remaining reasoning process conditioned on either the original question or the prefix-augmented input.}
    \label{fig:framework_comparison}
\end{figure*}

In standard RLVR frameworks,  the model is to generate a complete solution directly from the original prompt. Given a question \(q \in \mathcal{D}\), the model samples a reasoning trajectory \(r\) and is optimized according to the verifier reward:
\begin{equation}
    \theta^*
    =
    \operatorname*{argmax}_{\theta}
    \mathbb{E}_{q \sim \mathcal{D}, \; r \sim \pi_\theta(\cdot \mid q)}
    \big[ R(q, r) \big].
\label{eq:vanilla_grpo}
\end{equation}
However, since each reasoning step is conditioned only on the model's own previous generations, RLVR may reinforce a limited set of high-reward trajectories and leave alternative reasoning paths under-explored.

\subsection{Prefix-Completion RLVR}

To broaden exploration, we introduce prefix-completion RLVR. Instead of always generating a full solution from scratch, the target model is conditioned on a partial reasoning trajectory generated by an auxiliary model and trained to complete the remaining solution. Formally, let
\[
    \tilde{r} = \{\tilde{s}_1, \cdots, \tilde{s}_L\}
\]
denote an auxiliary prefix of \(L\) reasoning steps, and let \(r_{\mathrm{suf}}\) denote the continuation generated by the target model. The objective becomes
\begin{equation}
    \theta^*
    =
    \operatorname*{argmax}_{\theta}
    \mathbb{E}_{q \sim \mathcal{D}, \; r_{\mathrm{suf}} \sim \pi_\theta(\cdot \mid q, \tilde{r})}
    \big[ R(q, \tilde{r} \circ r_{\mathrm{suf}}) \big],
\label{eq:prefix_grpo}
\end{equation}
where \(\circ\) denotes concatenation.

By completing unfamiliar prefixes, the target model is encouraged to explore reasoning states outside its dominant generation distribution. Once a successful continuation receives a positive reward, RLVR can increase the likelihood of such alternative trajectories, thereby improving reasoning coverage after training.

\subsection{Entropy-Based Prefix Truncation}

To construct prefix trajectories, we use auxiliary small models to generate complete solutions and truncate them into partial reasoning prefixes. An effective prefix should perturb the target model away from its familiar reasoning patterns while leaving sufficient remaining steps for autonomous exploration. If the prefix is too long, the model may only output the final answer based on the complete trajectory in the prefix, while bypassing the intermediate reasoning process and hindering policy learning.

Based on the observation in Section~\ref{sec:cross_model_entropy}, cross-model prefixes induce high target-model entropy in the initial reasoning steps, but this entropy quickly decreases once the target model adapts to the auxiliary trajectory. We therefore identify the transition point by locating the largest entropy drop between adjacent reasoning steps. Specifically, after segmenting each auxiliary solution into steps, we compute the step-level entropy using the target base model and define
\begin{equation}
    L^*
    =
    \operatorname*{argmax}_{L }
    \left[
    \bar{H}_{\theta_0}(\tilde{s}_{L})
    -
    \bar{H}_{\theta_0}(\tilde{s}_{L+1})
    \right],
\end{equation}
where \(\theta_0\) denotes the parameters of the target base model before RLVR training. We then retain the prefix before the sharp entropy decrease:
\begin{equation}
    \tilde{r}
    =
    \{\tilde{s}_1, \cdots, \tilde{s}_{L^*}\}.
\end{equation}

This strategy avoids truncating in the middle of a reasoning step and preserves the high-entropy region that provides the strongest exploration signal. Since the step-level entropy trend remains largely stable during training, the truncation point is computed only once using the base model.

\subsection{Mixed Training}

Prefix-completion training promotes exploration, but the target model must still solve problems directly from the original question at inference time. We therefore adopt a mixed training strategy that combines standard question-only RLVR with prefix-completion RLVR.

For each prompt \(q\), with probability \(p\), we append the truncated auxiliary prefix \(\tilde{r}\) and train the model to complete the remaining reasoning process; with probability \(1-p\), we use the original question alone. The mixed objective is
\begin{equation*}
\begin{split}
    \theta^*
    =
    &\operatorname*{argmax}_{\theta}
    \mathbb{E}_{q \sim \mathcal{D}}
    \Big[
    (1-p)
    \mathbb{E}_{r \sim \pi_\theta(\cdot \mid q)}
    R(q,r)
    \\
    &+
    p
    \mathbb{E}_{r_{\mathrm{suf}} \sim \pi_\theta(\cdot \mid q,\tilde{r})}
    R(q,\tilde{r} \circ r_{\mathrm{suf}})
    \Big].
\end{split}
\end{equation*}

In our main experiments, we set \(p=0.2\), balancing exploration from prefix-completion examples with the model's ability to solve problems without external prefixes.

\section{Experiments}

\subsection{Training Setup}

\paragraph*{Model Selection}

We select small models of different model families from the target model, so that it's capable of generating solutions with diverse styles or different reasoning approaches, meanwhile not costing much computational resources. We choose Qwen2.5-7B \citep{qwen2.5} and Qwen2.5-Math-7B \citep{yang2024qwen25mathtechnicalreportmathematical} as the large models to train, and LLaMA-3.2-1B \citep{grattafiori2024llama} and Gemma-2-2B \citep{gemma_2024} as the small models to generate prefix. 

\paragraph*{Dataset and Evaluation}
For training, we employ GRPO algorithm \citep{shao2024deepseekmath}. We utilize MATH training set \citep{hendrycksmath2021}, which comprises 7500 unique problem-answer pairs. We test our model on various mathematical reasoning benchmarks, involving AIME 2024, AIME 2025, AMC 2023, MATH 500 \citep{hendrycksmath2021, lightman2024let}, Minerva \citep{lewkowycz2022solving} and Olympiad Bench \citep{he2024olympiadbench}, and compute the average pass@$k$ across the benchmarks. We adopt an unbiased estimation method to calculate pass@$k$, with $\text{pass@}k=1-\binom{N-c}{k}/\binom{N}{k}$, where $N$ denotes the total number of responses sampled per question, while $c$ denotes the number of correct ones. 

\paragraph*{Implementation Details}

We implement our training framework based on VeRL \citep{sheng2025hybridflow}. All the models are trained with learning rate of $10^{-6}$ and no warm-up steps. The training batch size is set to 1024 and PPO mini batch size is set to 256. Each input question is sampled with 8 responses with temperature 1.0. We don't employ the KL penalty in GRPO to stabilize the training process. For evaluation, we utilize vLLM \citep{kwon2023efficient} to accelerate generation. The temperature is set to 0.6, top p 0.95, with max generation length 4096. For prefix generation, we set temperature of small models to 0.4 to avoid lower data quality. For pass@$k$ calculation, we sample 128 responses per problem for MATH 500, Minerva and Olympiad Bench, and 200 responses for AIME 2024, AIME 2025 and AMC 2023 which are relatively small. 

\subsection{Main Results}

\begin{table*}[!htbp]
\centering
\renewcommand{\arraystretch}{1.3} 
\resizebox{\textwidth}{!}{
    \begin{tabular}{l *{8}{c}}
        \toprule
        \textbf{Method} & \textbf{pass@1} & \textbf{pass@2} & \textbf{pass@4} & \textbf{pass@8} & \textbf{pass@16} & \textbf{pass@32} & \textbf{pass@64} & \textbf{pass@128} \\
        \midrule
        Qwen2.5-7B & 27.83 & 36.37 & 44.07 & 51.01 & 57.33 & 62.99 & 67.98 & 72.15 \\
        Qwen2.5-7B + GRPO & 38.29 & 43.97 & 49.20 & 53.90 & 58.06 & 61.75 & 65.02 & 67.76 \\
        Qwen2.5-7B + LLaMA-3.2-1B prefix & 38.15 & 43.53 & 48.52 & 53.11 & 57.48 & 61.75 & 65.75 & 69.06 \\
        Qwen2.5-7B + Gemma-2-2B prefix & 39.01 & 44.60 & 49.52 & 54.03 & 58.41 & 62.66 & 66.78 & 70.71 \\
        \midrule[0.8pt]
        Qwen2.5-Math-7B & 26.45 & 36.83 & 46.22 & 53.77 & 59.58 & 64.29 & 68.52 & 72.71 \\
        Qwen2.5-Math-7B + GRPO & 40.08 & 45.86 & 51.13 & 55.91 & 60.19 & 63.72 & 66.63 & 69.30 \\
        Qwen2.5-Math-7B + Gemma-2-2B prefix & 40.11 & 45.98 & 51.32 & 56.35 & 61.04 & 65.06 & 68.26 & 71.22 \\
        \bottomrule
    \end{tabular}
}
\caption{Average pass@$k$ performance across all 6 mathematical reasoning benchmarks of different methods.}
\label{tab:pass_at_k_results}
\end{table*}


The main results are summarized in Table~\ref{tab:pass_at_k_results}. Incorporating prefixes generated by auxiliary small models consistently improves the performance of larger target models across all six reasoning benchmarks. Notably, the gains become more pronounced as \(k\) increases, indicating that our method effectively alleviates the narrowing of reasoning coverage caused by standard RLVR training.

Across different target models, prefixes generated by Gemma-2-2B generalize robustly to both Qwen2.5-7B and Qwen2.5-Math-7B. In comparison, prefixes generated by LLaMA-3.2-1B are particularly effective at preserving or improving pass@$128$ performance while maintaining comparable pass@$1$ accuracy. These results suggest that cross-model prefix guidance can improve exploration effectively.

\section{Analysis}

\begin{figure*}[t] 
    \centering
    \begin{subfigure}[b]{0.23\textwidth}
        \centering
        \includegraphics[width=\textwidth]{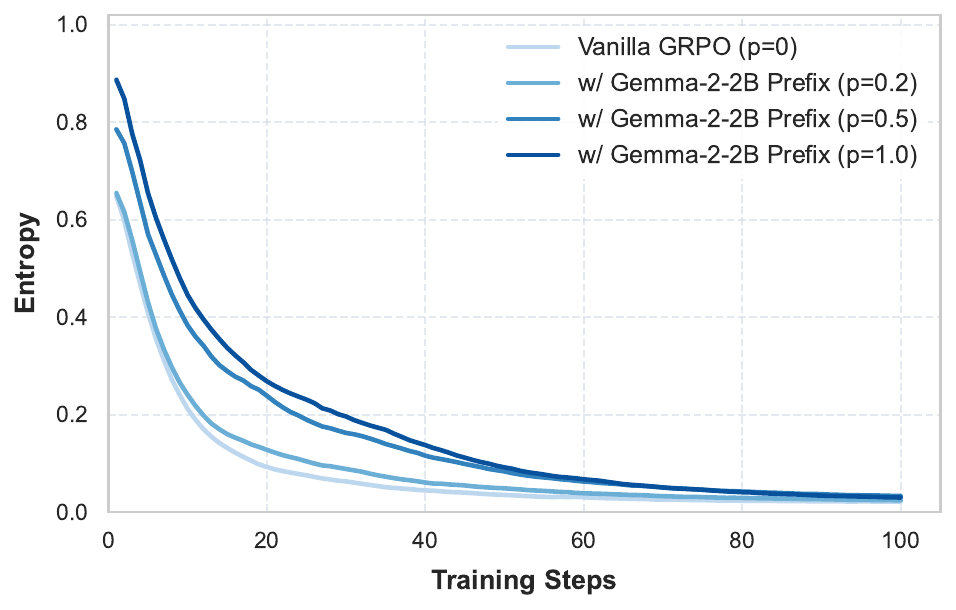}
        \caption{Policy Entropy}
        \label{fig:entropy}
    \end{subfigure}
    \hfill 
    \begin{subfigure}[b]{0.23\textwidth}
        \centering
        \includegraphics[width=\textwidth]{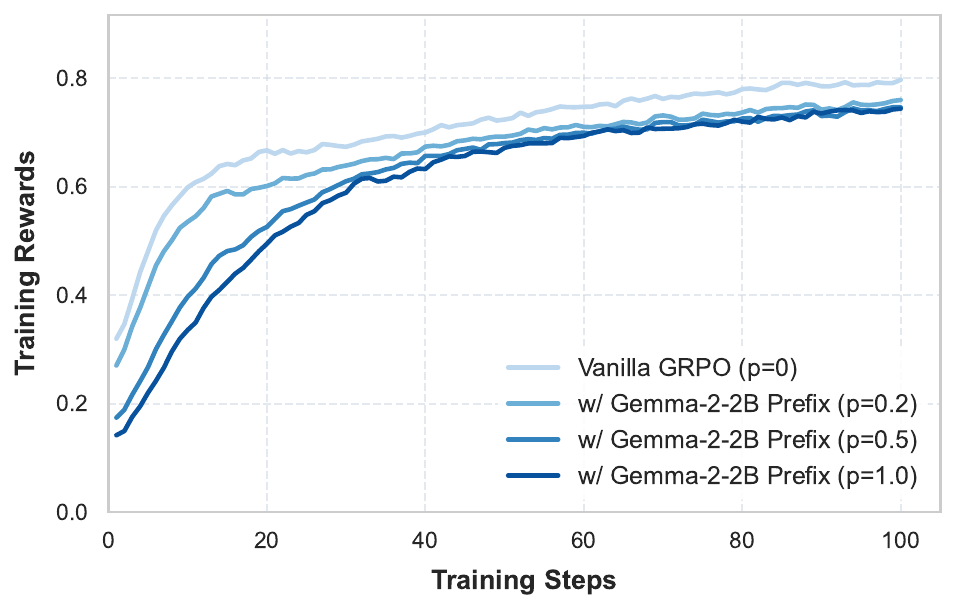}
        \caption{Average Rewards}
        \label{fig:rewards}
    \end{subfigure}
    \hfill
    \begin{subfigure}[b]{0.23\textwidth}
        \centering
        \includegraphics[width=\textwidth]{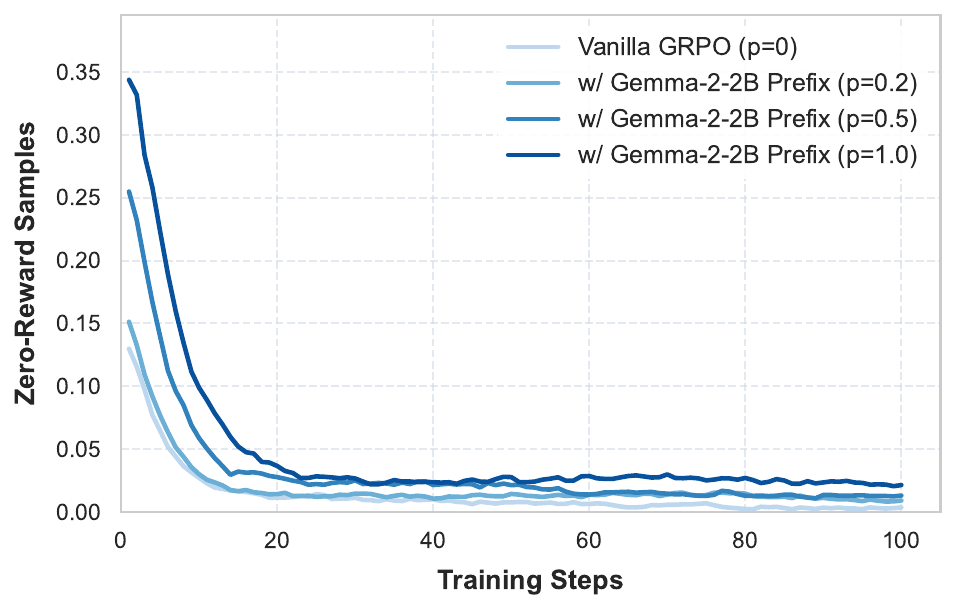}
        \caption{Zero-reward Ratio}
        \label{fig:zero_reward}
    \end{subfigure}
    \hfill
    \begin{subfigure}[b]{0.23\textwidth}
        \centering
        \includegraphics[width=\textwidth]{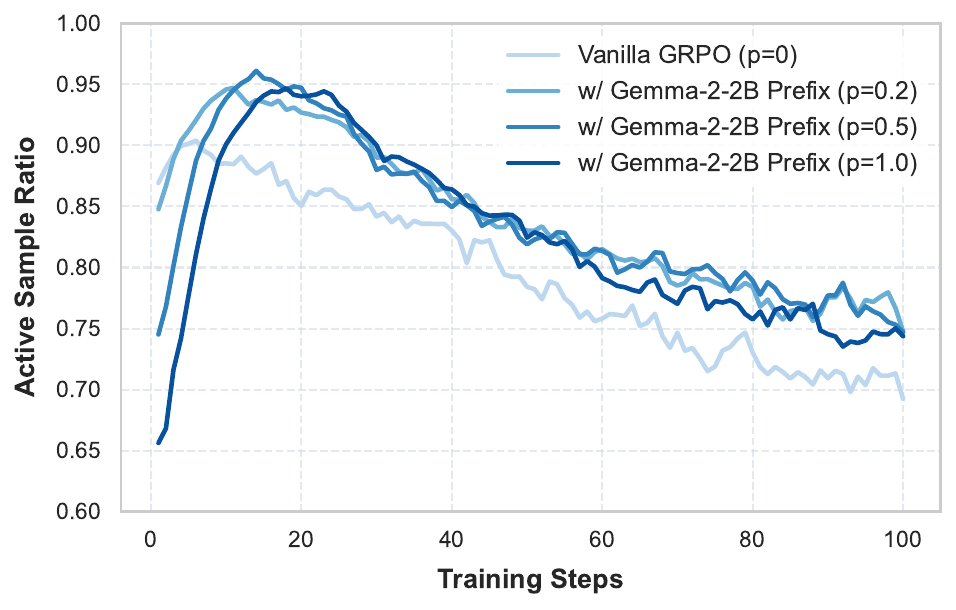}
        \caption{Non-trivial Sample Ratio}
        \label{fig:active_samples}
    \end{subfigure}
    
    \caption{Training dynamics under different prefix probabilities $p$.}
    \label{fig:training_dynamics}
\end{figure*}

\subsection{Training Dynamics}



Figure~\ref{fig:training_dynamics} compares the training dynamics of Qwen2.5-7B with/without Gemma-2-2B prefixes.

\paragraph*{Dynamic Exploration via Distributional Perturbation.}
As shown in Figure~\ref{fig:entropy}, our prefix-guided method maintains higher policy entropy than the vanilla GRPO baseline, especially in the early stage of training. Moreover, increasing the prefix injection probability leads to a consistent upward shift in the entropy trajectory. This suggests that heterogeneous prefixes serve as non-parametric steering signals, forcing the target model to continue from reasoning states that deviate from its own high-confidence trajectories.

Although the entropy eventually converges toward a level similar to the baseline, the early high-entropy phase is crucial. It delays premature policy collapse and gives the model more opportunities to explore alternative reasoning paths before the policy becomes overly concentrated. In this sense, prefix conditioning mitigates the over-confidence bias commonly observed in standard RLVR training.


\paragraph*{Exploration Cost and Informative Reward Signals.}
The lower average reward of prefix-guided models in Figure~\ref{fig:rewards} can be interpreted as the cost of exploration. Since the prefixes are generated by smaller auxiliary models, they are not always aligned with optimal solution paths and may even introduce misleading intermediate steps. However, as shown in Figure~\ref{fig:active_samples}, introducing these external prefixes effectively mitigates the reward signal sparsity inherent in GRPO.

In vanilla GRPO, samples within the same group often receive all-correct or all-incorrect responses, yielding negligible relative advantage signals. By introducing external prefixes, our method diversifies the sampled trajectories and produces more heterogeneous reward patterns. This provides more informative gradient signals during training, particularly in the early stages. In addition, Figure~\ref{fig:zero_reward} shows that although the target model may initially struggle to generate correct reasoning trajectory given prefixes from other models, it can eventually recover from imperfect prefixes and generate correct answers for most problems.

\begin{table}[htbp]
\centering
\renewcommand{\arraystretch}{1.2} 
\setlength{\tabcolsep}{6pt} 

\small 
\begin{tabular}{l *{2}{c}}
    \toprule
    \textbf{Method} & \textbf{pass@1} & \textbf{pass@128} \\
    \midrule
    $p=0$ (Vanilla GRPO) & 38.29 & 67.76 \\
    \midrule
    \multicolumn{3}{l}{\textit{Prefix Probability $p$}} \\
    \hspace{0.5em} $p=0.2$ & 39.01 & \textbf{70.71} \\
    \hspace{0.5em} $p=0.5$ & \textbf{39.22} & 67.89 \\
    \hspace{0.5em} $p=1.0$ & 38.56 & 68.54 \\
    \midrule
    \multicolumn{3}{l}{\textit{Truncation Strategy}} \\
    \hspace{0.5em} random step truncation & 38.08 & 70.19 \\
    \midrule
    \multicolumn{3}{l}{\textit{Homologous Prefix}} \\
    \hspace{0.5em} Qwen2.5-1.5B prefix & 38.88 & 67.22 \\
    \hspace{0.5em} Qwen2.5-7B prefix & 38.12 & 68.79 \\    
    \bottomrule
\end{tabular}

\caption{Ablation results on Qwen2.5-7B.}
\label{tab:ablation}
\end{table}
 
\subsection{Prefix Quality Assessment}

\begin{figure*}[t] 
    \centering
    \includegraphics[width=0.9\textwidth]{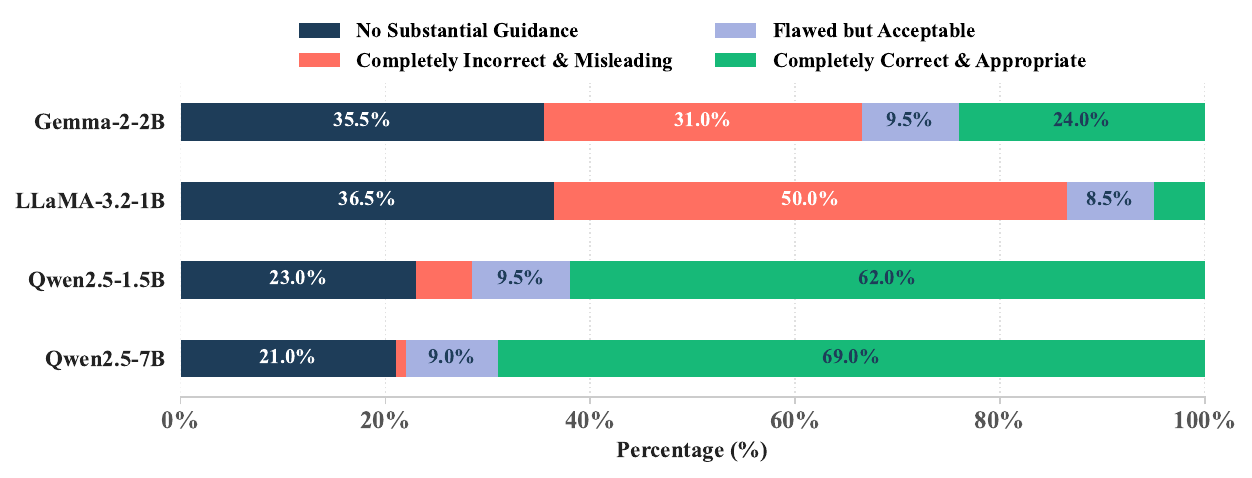}
    
    \caption{Fine-grained quality distribution of prefixes across different generator models.}
    
    \label{fig:overall_prefix_analysis}
\end{figure*}

To understand the underlying mechanism of our framework, we conducted an assessment on the prefix quality to see whether the initial reasoning steps provide inspiring or misleading guidance. 

We leverage DeepSeek-V4-Flash \citep{deepseekai2026deepseekv4} to measure the quality of the truncated prefixes generated by Gemma-2-2B, LLaMA-3.2-1B, Qwen2.5-1.5B and Qwen2.5-7B, respectively, and divide them into the following classes: (a) \textbf{No substantial guidance} (b) \textbf{Completely incorrect and misleading} (c) \textbf{Flawed but acceptable guidance} (d) \textbf{Completely correct with appropriate guidance}. We randomly pick 200 entries generated by each model, and employ DeepSeek to evaluate the prefixes based on a provided prompt template together with the original problems and ground truth answers. 


The results are illustrated in Figure~\ref{fig:overall_prefix_analysis}. Surprisingly, although prefix-completion training consistently improves performance by encouraging exploration, \textbf{the majority of prefixes generated by small models provide either no substantial guidance or even incorrect and misleading directions}. This suggests that the auxiliary small models do not function as expert teachers and are generally unable to provide high-fidelity distillation signals for the larger target model.

When combined with the observation that the proportion of zero-reward samples becomes negligible in the later stages of training, these results indicate that the target model can gradually recover from low-quality prefixes and still derive correct solutions within a limited number of samples. In other words, the benefit of prefix-completion training does not primarily come from the correctness of the prefix itself, but from its ability to expose the target model to unfamiliar reasoning states and stimulate broader exploration.


Additionally, among the non-homologous auxiliary models, Gemma-2-2B produces higher-fidelity prefixes than LLaMA-3.2-1B, which is consistent with its stronger empirical gains. In contrast, prefixes generated by Qwen2.5-1.5B and Qwen2.5-7B are of higher semantic quality and often provide more accurate intermediate guidance, yet they do not lead to comparable performance improvements. This is likely because these Qwen models are homologous to the target model and therefore generate reasoning trajectories that are distributionally closer to the target model's own outputs. As a result, their prefixes provide limited perturbation to the policy and fail to sufficiently expand the explored reasoning space.

This finding reveals a decoupling between semantic guidance accuracy and perturbation utility in RLVR. A useful prefix does not necessarily need to be the most correct or informative one; rather, it should move the target model away from its rigid, high-confidence reasoning trajectories. Therefore, cross-model prefixes can be viewed as a form of structured trajectory-level perturbation: they remain close to plausible reasoning processes, while being different enough from the target model's own generations to catalyze broader exploration.

\subsection{Ablation Study}

We further conduct a comprehensive ablation study to isolate the algorithmic contributions of our key components: the entropy-based truncation strategy, the prefix injection probability $p$, and the source of the prefix generator. The results are summarized in Table~\ref{tab:ablation}, from which we derive several crucial insights into the exploration dynamics.

First, our entropy-based truncation strategy  outperforms the random step-level truncation baseline. This empirical evidence indicates that initiating trajectories from states with high policy entropy effectively catalyzes productive exploration. 

Second, varying the injection probability $p$ unveils a delicate trade-off between exploration and exploitation; introducing excessive prefixes (${p=1.0}$) may lead to inconsistency between training and evaluation, thereby degrading downstream performance across all benchmarks. 

Lastly, replacing our smaller generator with a stronger but homologous model yields no noticeable gain compared to the vanilla GRPO baseline. The results imply the fact that the prefix completion paradigm works due to a perturbation in the reasoning trajectory, instead of a high-quality guidance.

\section{Conclusions}

We empirically study the potential impact of partial solutions generated by weaker models to the exploration of policy model in RLVR process. Though the outputs from weaker models tend to be incorrect or misleading, the elevated policy entropy further broaden the searching space of the target model, thus enhancing the pass@$k$ performance across various mathematical reasoning benchmarks. This work provides new insights into utilizing the small models and preserving the generative diversity during the RLVR training. 
Since our method operates strictly at the input-trajectory level, an exciting avenue for future work is integrating weak-model prefix guidance with advanced RLVR algorithms. We hypothesize that combining our data-level exploration with objective-level regularization could yield synergistic effects, further pushing the boundaries of LLM reasoning.

\section*{Limitations}

Though the proposed framework brings performance lift to RLVR training, the improvement is delicate with pre-configured hyper-parameters. Also, it currently only works on the mathematical reasoning tasks, while other domains like logic and code generations are under-explored.


\bibliography{custom}

\appendix

\section{Related Works}

\subsection{Reasoning with High-level Thought Guidance}
Standard methodologies utilize task decomposition \citep{zhou2022least, khot2022decomposed} or exemplar selection \citep{zhang2022automatic} to reduce problem complexity, but these strategies rely heavily on manual engineering and primarily enhance frozen models during inference \citep{yang2024buffer, wu2024beyond, yang2025reasonflux}. Effectively integrating abstract problem-solving guidance into on-policy RL training remains under-explored. While \citet{yan2026learning} introduces off-policy guidance to stabilize learning, it depends on strict distillation signals from a strong teacher where the capability gap often exacerbates training instability. In contrast, \citet{wu2025thought} integrates external high-level thought patterns directly into RL training, enhancing intrinsic reasoning and training stability without strong-policy dependencies.

\subsection{Exploration Mechanism and Learning Dynamics}
To resolve the exploration deficit in RLVR \citep{yue2025does, wu2025invisible}, current strategies manipulate inputs or objectives. Data-centric methods utilize off-policy distillation \citep{dong2025rl, li2025questa} or prompt paraphrasing \citep{liang2025beyond} to trigger diverse trajectories, while reward-centric methods optimize group-wise rewards to directly penalize redundancy \citep{walder2026pass, chen2025pass}. Alternatively, entropy-centric methods employ policy entropy \citep{cui2025entropy} or bounded token confidence via Token-Hidden Reward (THR) \citep{deng2025token} to control exploration, though THR incurs steep computational overhead. Additionally, \citet{peng2025simko} attempts to mitigate policy collapse via asymmetric gradient redistribution. It is worth noting that our prefix-completion framework is fundamentally different from these reward-centric or entropy-centric modifications. We do not alter the core GRPO algorithm or introduce heavy computational overhead. Instead, we provide a data-level perturbation strategy. Therefore, we view our approach not as a direct competitor to these algorithmic regularizations, but as a novel, lightweight paradigm that exposes the untapped potential of weaker models.

\section{Prompt Templates}
\label{sec:prompt}

We provide the prompt templates used for training and evaluation in our experiments. For 7B models namely Qwen2.5-7B and Qwen2.5-Math-7B, we are using the original chat template using SimpleRL \citep{zeng2025simplerl} style prompt in Figure~\ref{fig:7b_model_prompt}. For other smaller models involving Gemma-2-2B, LLaMA-3.2-1B and Qwen2.5-1.5B, we use prompt illustrated in Figure~\ref{fig:small_model_prompt} for high-quality prefix generation. For prefix quality assessment, we adopt a prompt template shown in Figure~\ref{fig:prefix_assessment_prompt}.

\begin{figure*}[htbp]
    \centering
    \begin{promptbox}{Prompt For 7B Models}
        \setlength{\parindent}{0pt}
        $<$|im\_start|$>$system \\
        You are a helpful assistant.$<$|im\_end|$>$ \\
        $<$|im\_start|$>$user \\
        \textcolor{q}{\{question\}}\\
        Please reason step by step, and put your final answer within \textbackslash{}boxed$\{\}$.
        $<$|im\_end|$>$ \\
        $<$|im\_start|$>$assistant
    \end{promptbox}
    \caption{Prompt for 7B models}
    \label{fig:7b_model_prompt}
    
\end{figure*}

\begin{figure*}[htbp]
    \centering
    \begin{promptbox}{Prompt For Smaller Models}
        \setlength{\parindent}{0pt}
        You are a helpful assistant.\\\\
        Question:\\\\
        \textcolor{q}{\{question\}}\\\\
        Solution:
    \end{promptbox}
    \caption{Prompt for smaller models}
    \label{fig:small_model_prompt}
\end{figure*}

\begin{figure*}[htbp]
    \centering
    \begin{promptbox}{Prompt For Prefix Assessment}
        \setlength{\parindent}{0pt}
        \textbf{System:}\\
        You are an expert mathematical evaluator. Your task is to analyze an incomplete reasoning prefix generated by a small model and evaluate the quality of guidance it provides for a target model to complete the problem. You must output your response strictly as a valid JSON object.\\\\Classification Criteria:\\1. 'Completely Correct \& Appropriate Guidance': The prefix logic and mathematical direction are entirely correct. It provides an optimal, flawless foundation that directly guides the next steps toward the correct answer.\\2. 'Flawed but Acceptable Guidance': The prefix contains minor computational slips, notation errors, or a slightly suboptimal route, but the core direction is reasonable and salvageable. It still provides a passable foundation.\\3. 'No Substantial Guidance': The prefix consists of trivial statements, repetitive filler text, or re-stating the prompt without making any real progress. It provides neither helpful direction nor harmful errors.\\4. 'Completely Incorrect \& Misleading Guidance': The prefix contains severe mathematical violations, fundamental logical collapses, or deep hallucinations. It actively misleads the reasoning path towards a dead-end or an incorrect answer.\\\\
        \textbf{User:}\\
        $<$input\_data$>$\\{[}Mathematical Problem{]}:\\\textcolor{q}{\{question\}}\\\\{[}Reference Answer \& Final Result{]}:\\\textcolor{q}{\{ground\_truth\_answer\}}\\\\{[}Small Model Reasoning Prefix (Incomplete Segment){]}:\\\textcolor{q}{\{small\_model\_prefix\}}\\$<$/input\_data$>$\\\\$<$output\_format$>$\\Return a JSON object with the exact following keys:\\\{\{\\  "category": "One of the four categories defined in the system prompt (exactly as stated).",\\  "prefix\_logic\_analysis": "A concise 1-2 sentence mathematical analysis of the prefix text (strictly under 60 words).",\\  "guidance\_impact\_justification": "Detailed explanation of how this specific level of guidance affects the downstream completion (e.g., whether it accelerates correct derivation, requires error-correction, wastes tokens, or derails the model)."\\\}\}\\$<$/output\_format$>$
        
    \end{promptbox}
    \caption{Prompt for DeepSeek-v4-Flash prefix assessment.}
    \label{fig:prefix_assessment_prompt}
\end{figure*}

\end{document}